\documentclass{article}
\usepackage{spconf,amsmath,epsfig}
\usepackage[colorlinks=true,linkcolor=NavyBlue,citecolor=BrickRed]{hyperref}
\usepackage[capitalise]{cleveref}
\usepackage{array}
\usepackage{url}
\usepackage{verbatim}
\usepackage[hang,flushmargin]{footmisc} 
\usepackage{graphicx}
\usepackage{multirow}
\usepackage{enumitem}
\usepackage{amssymb}
\usepackage[table,xcdraw,dvipsnames]{xcolor}
\usepackage{fancyhdr}
\fancypagestyle{copyright}{%
    \fancyhf{}
    \cfoot{\footnotesize%
        \textcopyright~2024 IEEE. Personal use of this material is permitted. Permission from IEEE must be obtained for all other uses, in any current or future media, including reprinting\slash republishing this material for advertising or promotional purposes, creating new collective works, for resale or redistribution to servers or lists, or reuse of any copyrighted component of this work in other works. This material is referenced by DOI: {\href{}{XXXXXXXXXX}}
    }
}

\title{AutoLCZ: Towards Automatized Local Climate Zone Mapping\\ from Rule-Based Remote Sensing}
\name{
  {Chenying Liu$^{1,2}$,\thanks{{The work of C. Liu, C. Albrecht} was funded by the Helmholtz Association through the Framework of \textit{HelmholtzAI}, grant ID: \texttt{ZT-I-PF-5-01} -- \textit{Local Unit Munich Unit @Aeronautics, Space and Transport (MASTr)}. The compute related to this work was supported by the Helmholtz Association's Initiative and Networking Fund on the HAICORE@FZJ partition.}}
  {Hunsoo Song$^{3}$,}
  {Anamika Shreevastava$^{4}$,}
  {Conrad M Albrecht$^{2}$}\thanks{%
    \textbf{CRediT}, \url{https://credit.niso.org}:
    \textit{Chenying Liu}: Investigation, Software, Visualization, Writing -- original draft;
    \textit{Hunsoo Song}: Methodology, Writing -- review \& editing;
    \textit{Anamika Shreevastava}: Methodology, Writing -- review \& editing, Supervision;
    \textit{Conrad M Albrecht} (corresponding author, \href{mailto:Conrad.Albrecht@DLR.de}{\tt Conrad.Albrecht@DLR.de}): Conceptualization, Methodology, Data curation, Resources, Writing -- review \& editing, Supervision
  }
}
\address{%
$^1$ Data Science in Earth Observation, Technical University of Munich, Germany\\
$^2$ Remote Sensing Technology Institute, German Aerospace Center, Germany\\
$^3$ Lyles School of Civil Engineering, Purdue University, USA\\
$^4$ Environmental Science and Engineering, Caltech, USA\\ 
}

\begin{document}
%
\maketitle

\begin{abstract}
Local climate zones (LCZs) established a standard classification system to categorize the landscape universe for improved urban climate studies.
Existing LCZ mapping is guided by human interaction with geographic information systems (GIS) or modelled from remote sensing (RS) data. GIS-based methods do not scale to large areas. However, RS-based methods leverage machine learning techniques to automatize LCZ classification from RS. Yet, RS-based methods require huge amounts of manual labels for training. 
We propose a novel LCZ mapping framework, termed \textit{AutoLCZ}, to extract the LCZ classification parameters from high-resolution RS modalities. We study the definition of numerical rules designed to mimic the LCZ definitions. Those rules model geometric and surface cover parameters from LiDAR data. Correspondingly, we enable LCZ classification from RS data in a GIS-based scheme. The proposed \textit{AutoLCZ} method has potential to reduce the human labor to acquire accurate metadata. At the same time, \textit{AutoLCZ} sheds light on the physical interpretability of RS-based methods. In a proof-of-concept for New York City (NYC) we leverage airborne LiDAR surveys to model four LCZ parameters to distinguish eight LCZ types. The results indicate the potential of \textit{AutoLCZ} as a promising avenue for large-scale LCZ mapping from RS data.
\end{abstract}
\thispagestyle{copyright}
\pagestyle{plain}

\section{Introduction}
\label{sec:intro}

\begin{table*}[t]
\centering
\caption{LCZ parameters and corresponding parameter ranges to distinguish 10 built types. The geometric and surface cover parameters carry no dimension, except for the height of roughness elements (in meters). All values are representative of the local scale. The parameters considered in this work are highlighted by a blue background. Reference \cite{stewart_local_2012} contains details.}
\label{tab:properties}
\scriptsize
\begin{tabular}
{m{1.cm}<{\centering}|m{2.8cm}<{\centering}|m{1.cm}<{\centering}m{1.cm}<{\centering}m{1.cm}<{\centering}m{1.cm}<{\centering}m{1.cm}<{\centering}m{1.cm}<{\centering}m{1.cm}<{\centering}|m{1.cm}<{\centering}m{1.cm}<{\centering}m{1.cm}<{\centering}}
\hline\hline
                          &                              & \multicolumn{7}{c|}{Geometric and surface cover parameters}                                                                                                                                                                                                                                                                              & \multicolumn{3}{c}{Thermal, radiative, and metabolic parameters}                     \\ \cline{3-12} 
\multirow{-2}{*}{No. LCZ} & \multirow{-2}{*}{Built type} & \multicolumn{1}{c}{SVF$^a$}           & \multicolumn{1}{c}{AR$^b$}          & \multicolumn{1}{c}{\cellcolor[HTML]{ECF4FF}BSF$^c$} & \multicolumn{1}{c}{\cellcolor[HTML]{ECF4FF}ISF$^d$}       & \multicolumn{1}{c}{\cellcolor[HTML]{ECF4FF}PSF$^e$}       & \multicolumn{1}{c}{\cellcolor[HTML]{ECF4FF}HRE$^f$}          & TRC$^g$ & \multicolumn{1}{c}{SAD$^h$}     & \multicolumn{1}{c}{SAL$^i$}   & AHO$^j$           \\ 
\hline
1                         & Compact high-rise            & \multicolumn{1}{c}{0.2-0.4}           & \multicolumn{1}{c}{\textgreater{}2} & \multicolumn{1}{c}{\cellcolor[HTML]{ECF4FF}40-60}   & \multicolumn{1}{c}{\cellcolor[HTML]{ECF4FF}40-60}         & \multicolumn{1}{c}{\cellcolor[HTML]{ECF4FF}\textless{}10} & \multicolumn{1}{c}{\cellcolor[HTML]{ECF4FF}\textgreater{}25} & 8       & \multicolumn{1}{c}{1,500-1,800} & \multicolumn{1}{c}{0.10-0.20} & 50-300            \\ 
2                         & Compact midrise              & \multicolumn{1}{c}{0.3-0.6}           & \multicolumn{1}{c}{0.75-2}          & \multicolumn{1}{c}{\cellcolor[HTML]{ECF4FF}40-70}   & \multicolumn{1}{c}{\cellcolor[HTML]{ECF4FF}30-50}         & \multicolumn{1}{c}{\cellcolor[HTML]{ECF4FF}\textless{}20} & \multicolumn{1}{c}{\cellcolor[HTML]{ECF4FF}10-25}            & 6-7     & \multicolumn{1}{c}{1,500-2,200} & \multicolumn{1}{c}{0.10-0.20} & \textless{}75     \\ 
3                         & Compact low-rise             & \multicolumn{1}{c}{0.2-0.6}           & \multicolumn{1}{c}{0.75-1.5}        & \multicolumn{1}{c}{\cellcolor[HTML]{ECF4FF}40-70}   & \multicolumn{1}{c}{\cellcolor[HTML]{ECF4FF}20-50}         & \multicolumn{1}{c}{\cellcolor[HTML]{ECF4FF}\textless{}30} & \multicolumn{1}{c}{\cellcolor[HTML]{ECF4FF}3-10}             & 6       & \multicolumn{1}{c}{1,200-1,800} & \multicolumn{1}{c}{0.10-0.20} & \textless{}75     \\ 
4                         & Open high-rise               & \multicolumn{1}{c}{0.5-0.7}           & \multicolumn{1}{c}{0.75-1.25}       & \multicolumn{1}{c}{\cellcolor[HTML]{ECF4FF}20-40}   & \multicolumn{1}{c}{\cellcolor[HTML]{ECF4FF}30-40}         & \multicolumn{1}{c}{\cellcolor[HTML]{ECF4FF}30-40}         & \multicolumn{1}{c}{\cellcolor[HTML]{ECF4FF}\textgreater{}25} & 7-8     & \multicolumn{1}{c}{1,400-1,800} & \multicolumn{1}{c}{0.12-0.25} & \textless{}50     \\ 
5                         & Open midrise                 & \multicolumn{1}{c}{0.5-0.8}           & \multicolumn{1}{c}{0.3-0.75}        & \multicolumn{1}{c}{\cellcolor[HTML]{ECF4FF}20-40}   & \multicolumn{1}{c}{\cellcolor[HTML]{ECF4FF}30-50}         & \multicolumn{1}{c}{\cellcolor[HTML]{ECF4FF}20-40}         & \multicolumn{1}{c}{\cellcolor[HTML]{ECF4FF}10-25}            & 5-6     & \multicolumn{1}{c}{1,400-2,000} & \multicolumn{1}{c}{0.12-0.25} & \textless{}25     \\ 
6                         & Open low-rise                & \multicolumn{1}{c}{0.6-0.9}           & \multicolumn{1}{c}{0.3-0.75}        & \multicolumn{1}{c}{\cellcolor[HTML]{ECF4FF}20-40}   & \multicolumn{1}{c}{\cellcolor[HTML]{ECF4FF}20-50}         & \multicolumn{1}{c}{\cellcolor[HTML]{ECF4FF}30-60}         & \multicolumn{1}{c}{\cellcolor[HTML]{ECF4FF}3-10}             & 5-6     & \multicolumn{1}{c}{1,200-1,800} & \multicolumn{1}{c}{0.12-0.25} & \textless{}25     \\ 
7                         & Lightweight low-rise         & \multicolumn{1}{c}{0.2-0.5}           & \multicolumn{1}{c}{1-2}             & \multicolumn{1}{c}{\cellcolor[HTML]{ECF4FF}60-90}   & \multicolumn{1}{c}{\cellcolor[HTML]{ECF4FF}\textless{}20} & \multicolumn{1}{c}{\cellcolor[HTML]{ECF4FF}\textless{}30} & \multicolumn{1}{c}{\cellcolor[HTML]{ECF4FF}2-4}              & 4-5     & \multicolumn{1}{c}{800-1,500}   & \multicolumn{1}{c}{0.15-0.35} & \textless{}35     \\ 
8                         & Large low-rise               & \multicolumn{1}{c}{\textgreater{}0.7} & \multicolumn{1}{c}{0.1-0.3}         & \multicolumn{1}{c}{\cellcolor[HTML]{ECF4FF}30-50}   & \multicolumn{1}{c}{\cellcolor[HTML]{ECF4FF}40-50}         & \multicolumn{1}{c}{\cellcolor[HTML]{ECF4FF}\textless{}20} & \multicolumn{1}{c}{\cellcolor[HTML]{ECF4FF}3-10}             & 5       & \multicolumn{1}{c}{1,200-1,800} & \multicolumn{1}{c}{0.15-0.25} & \textless{}50     \\
9                         & Sparsely built               & \multicolumn{1}{c}{\textgreater{}0.8} & \multicolumn{1}{c}{0.1-0.25}        & \multicolumn{1}{c}{\cellcolor[HTML]{ECF4FF}10-20}   & \multicolumn{1}{c}{\cellcolor[HTML]{ECF4FF}\textless{}20} & \multicolumn{1}{c}{\cellcolor[HTML]{ECF4FF}60-80}         & \multicolumn{1}{c}{\cellcolor[HTML]{ECF4FF}3-10}             & 5-6     & \multicolumn{1}{c}{1,000-1,800} & \multicolumn{1}{c}{0.12-0.25} & \textless{}10     \\ 
10                        & Heavy industry               & \multicolumn{1}{c}{0.6-0.9}           & \multicolumn{1}{c}{0.2-0.5}         & \multicolumn{1}{c}{\cellcolor[HTML]{ECF4FF}20-30}   & \multicolumn{1}{c}{\cellcolor[HTML]{ECF4FF}20-40}         & \multicolumn{1}{c}{\cellcolor[HTML]{ECF4FF}40-50}         & \multicolumn{1}{c}{\cellcolor[HTML]{ECF4FF}5-15}             & 5-6     & \multicolumn{1}{c}{1,000-2,500} & \multicolumn{1}{c}{0.12-0.20} & \textgreater{}300 \\ 
\hline\hline
\end{tabular}

\tiny{$^a$Sky View Factor (SVF): ratio of the amount of sky hemisphere visible from ground level to that of an unobstructed hemisphere;
$^b$Aspect ratio (AR): mean height-to-width ratio of street canyons (LCZs 1-7), and building spacing (LCZs 8-10);
$^c$Building Surface Fraction (BSF): ratio of building area to total area; 
$^d$Impervious Surface Fraction (ISF): ratio of impervious area to total area; 
$^e$Pervious Surface Fraction (PSF): ratio of pervious area to total area; \\
$^f$Height of Roughness Elements (HRE): geometric average of building heights;
$^g$Terrain Roughness Class (TRC): classification of effective terrain roughness for city and country landscapes; \\
$^h$Surface ADmittance (SAD); 
$^i$Surface ALbedo (SAL);
$^j$Anthropogenic Heat Output (AHO)}
\end{table*}

\begin{figure*}
    \centering
    \begin{tabular}{m{2.6cm}<{\centering}m{2.6cm}<{\centering}m{2.8cm}<{\centering}m{2.6cm}<{\centering}m{2.6cm}<{\centering}m{0.8cm}<{\centering}}
         \includegraphics[height=2.7cm]{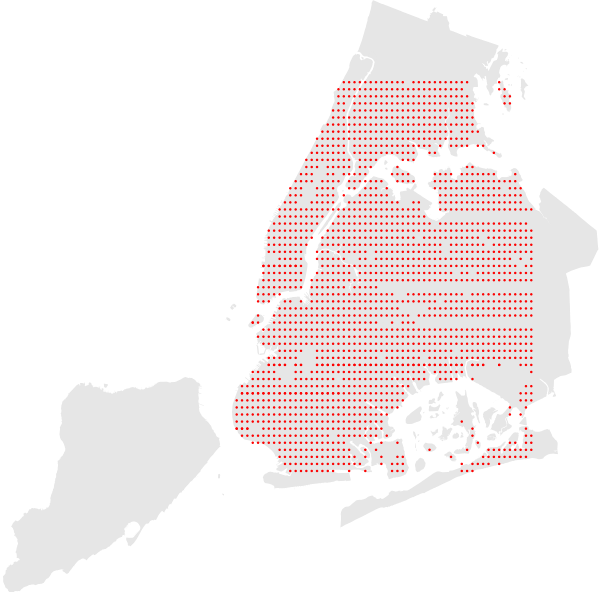} &
         \includegraphics[height=2.7cm]{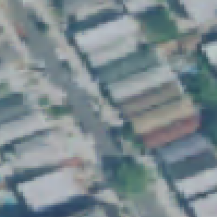} & 
         \includegraphics[height=2.7cm]{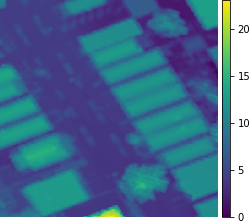} &
         \includegraphics[height=2.7cm]{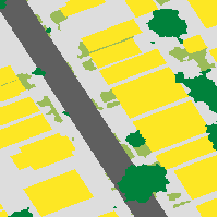} &
         \includegraphics[height=2.7cm]{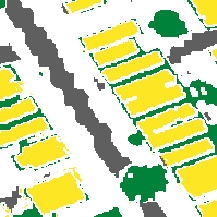} &
         \includegraphics[height=2.6cm]{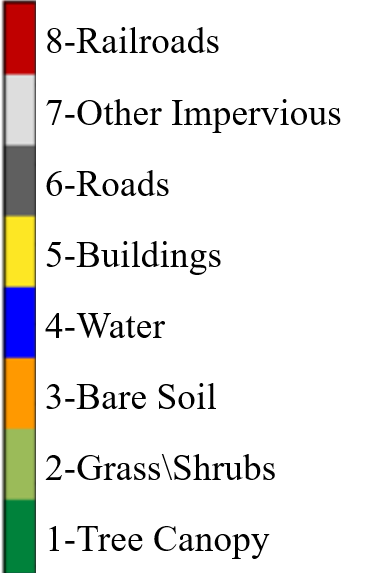}\\
         (a) sampling areas & (b) optical & (c) elevation (m) & (d) ground truth & (e) noisy labels
    \end{tabular}
    \caption{Overview on the data of our study: (a) sampling areas in NYC boroughs Manhattan, Queens, Bronx, and Brooklyn, where each point represents a 0.64$^2$ km$^2$ squared region. Illustrations of a built type \textit{compact low-rise} (LCZ 3) from (b) the optical view, (c) the 2D mean elevation statistics rasterized from LiDAR data, (d) the ground truth land cover mask, and (e) the noisy land cover mask derived from LiDAR statistics  (cf.\ \cite{albrecht_autogeolabel_2021}).}
    \label{fig:data}
\end{figure*}

The concept of Local Climate Zones (LCZs) was established by Stewart and Oke to facilitate metadata communication in urban temperature observations and heat island studies \cite{stewart_local_2012}. The LCZ classification system defines 17 classes, including 10 built types and 7 land cover types to characterize constructed sites and natural scenes. These classes are locally defined based on climate-relevant surface properties, encompassing surface cover types, structures, materials, and human activities.
The correlation between local climate conditions and spatial physical factors, such as building height and spacing, pervious surface fraction, tree density, and soil wetness, has been substantiated by subsequent researches \cite{stewart_evaluation_2014}. This evidence supports the LCZ partitions, establishing it as a useful tool widely applied in various urban climate related studies \cite{kotharkar_evaluating_2018, bechtel_suhi_2019,10281573}.

In recent years, many attempts have been made to generate more accurate LCZ maps. The existing LCZ classification methods can be roughly grouped into three categories, that is, geographic information system (GIS)-based, remote sensing (RS)-based, and combined methods \cite{huang_mapping_2023}.
GIS-based methods follow the instructions proposed in \cite{stewart_local_2012}, choosing and collecting appropriate site metadata to quantify the defined parameters for LCZ classification. These methods have a good physical interpretability, yet require high-quality data to model accurate LCZ parameters to guarantee the reliability of derived results \cite{zheng_gis-based_2018}. Such data are hard to obtain especially for large-scale studies. Besides, metadata sometimes cannot be perfectly matched with given value ranges due to the heterogeneity of local surface structures in real scenarios. Subclasses are often introduced to encompass these exceptions in the specific study area. 
On the contrary, RS-based methods utilize machine learning techniques to classify LCZs from remote sensing imagery in a supervised fashion, promising in producing large-scale LCZ maps. These methods heavily rely on training samples to explore the projection from RS image features to LCZs, which ignores underlining physical characteristics \cite{qiu_local_2019}. Accuracy assessment is necessary to evaluate the learned projections. In this case, training labels are usually collected by human visually inspecting local scenes. This approach easily leads to misclassified labels partly due to the vague definitions of LCZ classes \cite{zhu_so2sat_2020}. 
Combined methods aim to couple the two kinds of methods by serialization \cite{gal_comparison_2015} or constructing compound feature sets \cite{zhou_mapping_2020}, which integrate not only advantages but also disadvantages.

In this work, we propose a new combined LCZ mapping framework called \textit{AutoLCZ}, which learns the projections from the estimated LCZ parameters with the aid of rules that scalably operate on high-resolution, RS modalities, rather than from original RS features. Among others, large-scale airborne LiDAR surveys have been made available in recent years. LiDAR (Light Detection and Ranging) generates 3D data points from laser pulses bouncing back from the Earth’s surface to measure ranges\slash distances, which is the major data source to map elevation and topography. The rich structural information provided by LiDAR data points is also useful in uncovering land cover characteristics such as trees, roads, buildings, etc.\ \cite{albrecht_autogeolabel_2021}. These aspects render LiDAR surveys suitable for modeling the geometric and land cover parameters for LCZ classification. 
We explore our approach in a proof-of-concept for the City of New York (NYC), where we only consider the eight built types out of ten excluding LCZs 7 and 9, given the fact that NYC is a highly urbanized area without lightweight low-rise or sparsely built areas. As illustrated in \cref{tab:properties}, we model four geometric and surface cover parameters in this work as a first step to test the feasibility of the proposed framework.

In the following, we will first give a brief introduction of the study area along with related data in \cref{sec:data}. Then, \cref{sec:meth} presents the details of \textit{AutoLCZ}, followed by the experimental results in \cref{sec:exp}, and conclusions for future lines in \cref{sec:conc}.

\section{Study area \& data}  \label{sec:data}

We selected NYC as our study area for its coverage by various data modalities, such as: LiDAR data, high and moderate resolution optical data, land cover masks, and LCZ masks. They are open sourced thanks to the government's efforts of boosting data sharing\footnote{e.g., available at \url{https://opendata.cityofnewyork.us}}. As illustrated in \cref{fig:data}a, we sampled 2390 regions of size 0.64$^2$ km$^2$ across the study area. We subsequently cropped them into 34,920 small patches of 100m$\times$100m after data cleansing. The optical image sample in \cref{fig:data}b corresponds to a National Agriculture Imaging Program’s (NAIP) orthophoto. The majority of data used in this work derives from LiDAR statistics and the ground truth land cover masks.

\textbf{LiDAR surveys}: Point cloud LiDAR data were originally acquired in 2017 with an approximate density of 10 points per square meter. We converted the raw 3D point clouds into raster layers using a 1.5m diameter sliding circle with a grid size of 0.5m to accumulate statistics such as minimum, maximum, mean, and standard deviation of each attribute. In LCZ parameter modeling, we mainly utilize the elevation-mean statistics as in \cref{fig:data}c.

\textbf{Land cover masks}: NYC agencies made the 8-class land cover masks publicly available with the aid of the 2017 NYC LiDAR data and additional geospatial surveys like building footprints, and overhead imagery. Such detailed land use information is crucial for LCZ mapping, yet rare to come by and costly to process. As an alternative, we explore the potential of noisy land cover masks as in \cref{fig:data}e, which are derived from LiDAR statistics by \textit{AutoGeoLabel} \cite{albrecht_autogeolabel_2021} towards further advancing automation of LCZ mapping. Only three specific classes, including trees, buildings, and roads, are delineated in this way. The rest remains unlabeled (white in \cref{fig:data}e, \textit{background}). The effectiveness of these rough labels have been verified in real applications such as urban forest monitoring \cite{albrecht_monitoring_2022}. We will show in the following that these easy-to-come-by labels, though less sufficient in describing the surface than ground truth masks, are still useful in LCZ mapping.

\section{Methodology} \label{sec:meth}

\subsection{LCZ Parameter Modeling} \label{sec:meth:param}

Let $\mathbf{D}=\{(\mathbf{h}^{(i)}, \mathbf{y}^{(i)}, z^{(i)})\vert~i=1,\dots,N\}=\{(\mathbf h, \mathbf y, z)\}_N$ denote the dataset consisting of $N=34,920$ image patches with LiDAR mean-elevation statistics $\mathbf{h}^{(i)} \in {\mathbb{R}}^{n\times n}$ and land cover masks $\mathbf{y}^{(i)} \in {\mathbb{L}}^{n\times n}$ labelled by LCZ classes $z^{(i)}\in\mathbb{Z}$, where $\mathbb{R}$ is the set of real numbers, $\mathbb{L}=\{{1,\cdots,L}\}$ denotes the category set of land cover masks, $\mathbb{Z}=\{1,\cdots,Z\}$ entails the class definitions of LCZ, and $n$ represents the linear size of patches with $n^2$ pixels---the unit size for LCZ mapping. For the NYC dataset, we have $L=8$ as in \cref{fig:data}, $Z=10$, $n=200$ with the spatial resolution of $z^{(i)}$ being 100m and that of $\mathbf{h}^{(i)}$ and $\mathbf{y}^{(i)}$ being 0.5m. Both ground truth land cover masks $\mathbf{y}^{(i)}$ and noisy land cover masks $\mathbf{y}^{(i)}=\mathbf{y}^{(i)}\left(\mathbf h^{(i)}\right)$ serve as label data for our experiments in \cref{sec:exp}.

The three surface fraction factors
$\text{SF}\in\{\text{BSF}, \text{ISF}, \text{PSF}\}$ we derive from $\mathbf{y}^{(i)}$ as follows:
\begin{equation}
    \label{eq:SF}
    \text{SF}^{(i)}=\frac{1}{n^2}\sum_{l\in\mathbb{L}_\text{SF}}\sum_{p,q=1}^n\delta\left(y^{(i)}_{pq}-l\right)
\end{equation}
where $\delta(x)=1$ if $x=0$, otherwise 0. $l$ is the target class index representing: building, impervious, and pervious surfaces for BSF, ISF, and PSF, respectively. For the NYC dataset: $\mathbb{L}_\text{BSF}=\{5\}$, $\mathbb{L}_\text{ISF}=\{6,7\}$, and $\mathbb{L}_\text{PSF}=\{1,2,3\}$, according to \cref{fig:data}.
The Height of Roughness Elements (HRE) is estimated by the geometric mean of heights over building pixels in $\mathbf{h}^{(i)}$, that is: 
\begin{align}
    \label{eq:HRE}
    \log\text{HRE}^{(i)}&=\frac{1}{m^{(i)}}\sum_{p,q=1}^n\delta\left(y^{(i)}_{pq}-5\right)\log h^{(i)}_{pq}\nonumber\\
    m^{(i)}&=\sum_{p,q=1}^n\delta\left(y^{(i)}_{pq}-5\right)
\end{align}
for the $i$th image patch such that $\text{HRE}=\exp \langle\log\mathbf h\rangle$. The symbol $\langle\cdot\rangle$ denotes arithmetic averaging over building areas computing the logarithm of the mean-elevation statistics $\mathbf h$.

\subsection{LCZ Classification} \label{sec:meth:cls}
\label{sec:LCZClass}
Since the four LCZ parameters BSF, ISF, PSF, and HRE are estimated following the initial definition of the LCZs \cite{stewart_evaluation_2014}, a naive approach to obtain LCZ classification results applies the thresholds listed in \cref{tab:properties}. However, the estimation bias caused by data noise and inaccurate approximation of rules likely leads to unsatisfying results. Thus, we recompute the thresholds in a data-driven fashion from the LCZ labels $z$. Our approach is based on the ground truth land cover masks $\mathbf y$ and the mean-elevation data $\mathbf h$. Specifically, within each class $\mathbb L$, we calculate the mean $\mu$ and standard deviation $\sigma$ for each of the SFs and HRE employing \cref{eq:SF,eq:HRE}. Following the assumption that about 95\% of Gaussian noise--distributed values fall into the range of $\Delta=[\mu-2\sigma,\mu+2\sigma]$ \cite{noauthor_two-sigma_2008}, we set the lower and upper bounds for each LCZ parameter as $\mu-2\sigma$ and $\mu+2\sigma$, respectively. 

\section{Experiments} \label{sec:exp}

\subsection{Estimated LCZ Parameter Thresholds} \label{sec:exp:thr}

For the SFs and HRE parameters, \cref{tab:thr} summarizes the estimated thresholds as per \cref{sec:LCZClass} along with their definition in \cite{stewart_local_2012}. We report the results of 8 classes due to the absence of classes 7 and 9 from the NYC dataset. According to \cref{tab:thr}, most of the LCZ definitions deviate from the statistics such that the estimated thresholds exhibit broader value ranges. Besides the limits of assuming  Gaussians distributions for the SFs and HRE parameters per LCZ class, noise in the LCZ labels restricts the overall accuracy due to visual interpretation by human inspection.

\begin{table}[t]
\scriptsize
\centering
\caption{Intervals $\Delta_j^{(\hat z)}$ of parameters $x_j$ to predict LCZs $\hat z$: (a) by definition (black, min--max or $<$max or $>$min), and (b) by estimated thresholds from LCZ labels $z$ and ground truth land cover masks $\mathbf y$ (blue, $\Delta$=[min, max]).}
\label{tab:thr}
\begin{tabular}{c|cccc}
\hline\hline
$\hat z=\text{LCZ}$     & $x_1=\text{BSF}$                                    & $x_2=\text{ISF}$                                    & $x_3=\text{PSF}$                                    & $x_4=\text{HRE}$                                      \\
\hline\hline
                     & 0.4-0.6                                         & 0.4-0.6                                         & \textless{}0.1                                  & {\color[HTML]{333333} \textgreater{}25}           \\
\multirow{-2}{*}{1}  & {\color[HTML]{3531FF} \textbf{{[}0.30,~0.65{]}}} & {\color[HTML]{3531FF} \textbf{{[}0.25,~0.55{]}}} & {\color[HTML]{3531FF} \textbf{{[}0.00,~0.26{]}}} & {\color[HTML]{3531FF} \textbf{{[}18.98,~69.14{]}}} \\
\hline
                     & 0.4-0.7                                         & 0.3-0.5                                         & \textless{}0.2                                  & 10-25                                             \\
\multirow{-2}{*}{2}  & {\color[HTML]{3531FF} \textbf{{[}0.18,~0.50{]}}} & {\color[HTML]{3531FF} \textbf{{[}0.31,~0.62{]}}} & {\color[HTML]{3531FF} \textbf{{[}0.04,~0.32{]}}} & {\color[HTML]{3531FF} \textbf{{[}4.81,~24.25{]}}}  \\
\hline
                     & 0.4-0.7                                         & 0.2-0.5                                         & \textless{}0.3                                  & 3-10                                              \\
\multirow{-2}{*}{3}  & {\color[HTML]{3531FF} \textbf{{[}0.22,~0.42{]}}} & {\color[HTML]{3531FF} \textbf{{[}0.35,~0.58{]}}} & {\color[HTML]{3531FF} \textbf{{[}0.09,~0.32{]}}} & {\color[HTML]{3531FF} \textbf{{[}4.20,~17.72{]}}}  \\
\hline
                     & 0.2-0.4                                         & 0.3-0.4                                         & 0.3-0.4                                         & \textgreater{}25                                  \\
\multirow{-2}{*}{4}  & {\color[HTML]{3531FF} \textbf{{[}0.05,~0.34{]}}} & {\color[HTML]{3531FF} \textbf{{[}0.21,~0.58{]}}} & {\color[HTML]{3531FF} \textbf{{[}0.17,~0.61{]}}} & {\color[HTML]{3531FF} \textbf{{[}2.20,~29.78{]}}}  \\
\hline
                     & 0.2-0.4                                         & 0.3-0.5                                         & 0.2-0.4                                         & 10-25                                             \\
\multirow{-2}{*}{5}  & {\color[HTML]{3531FF} \textbf{{[}0.11,~0.37{]}}} & {\color[HTML]{3531FF} \textbf{{[}0.22,~0.53{]}}} & {\color[HTML]{3531FF} \textbf{{[}0.21,~0.54{]}}} & {\color[HTML]{3531FF} \textbf{{[}6.44,~25.09{]}}}  \\
\hline
                     & 0.2-0.4                                         & 0.2-0.5                                         & 0.3-0.6                                         & 3-10                                              \\
\multirow{-2}{*}{6}  & {\color[HTML]{3531FF} \textbf{{[}0.04,~0.28{]}}} & {\color[HTML]{3531FF} \textbf{{[}0.20,~0.56{]}}} & {\color[HTML]{3531FF} \textbf{{[}0.23,~0.68{]}}} & {\color[HTML]{3531FF} \textbf{{[}0.09,~18.11{]}}}  \\
\hline
                     & 0.3-0.5                                         & 0.4-0.5                                         & \textless{}0.2                                  & 3-10                                              \\
\multirow{-2}{*}{8}  & {\color[HTML]{3531FF} \textbf{{[}0.04,~0.59{]}}} & {\color[HTML]{3531FF} \textbf{{[}0.31,~0.81{]}}} & {\color[HTML]{3531FF} \textbf{{[}0.00,~0.27{]}}} & {\color[HTML]{3531FF} \textbf{{[}3.25,~12.21{]}}}  \\
\hline
                     & 0.2-0.3                                         & 0.2-0.4                                         & 0.4-0.5                                         & 5-15                                              \\
\multirow{-2}{*}{10} & {\color[HTML]{3531FF} \textbf{{[}0.03,~0.49{]}}} & {\color[HTML]{3531FF} \textbf{{[}0.32,~0.81{]}}} & {\color[HTML]{3531FF} \textbf{{[}0.00,~0.30{]}}} & {\color[HTML]{3531FF} \textbf{{[}2.59,~14.16{]}}} \\
\hline\hline
\end{tabular}
\end{table}

\begin{figure*}
    \centering
    \begin{tabular}{m{2.8cm}<{\centering}m{3.1cm}<{\centering}m{2.8cm}<{\centering}m{2.8cm}<{\centering}m{3.7cm}<{\centering}}
         \includegraphics[height=2.7cm]{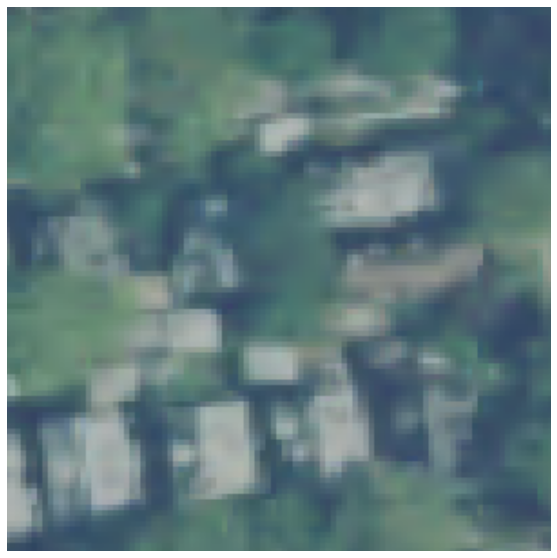} &
         \includegraphics[height=2.7cm]{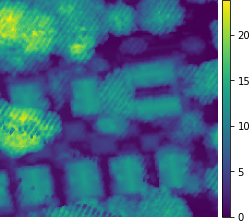} &
         \includegraphics[height=2.7cm]{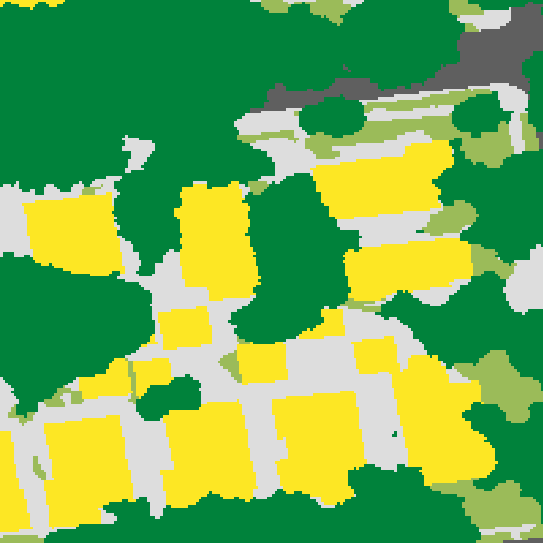} &
         \includegraphics[height=2.7cm]{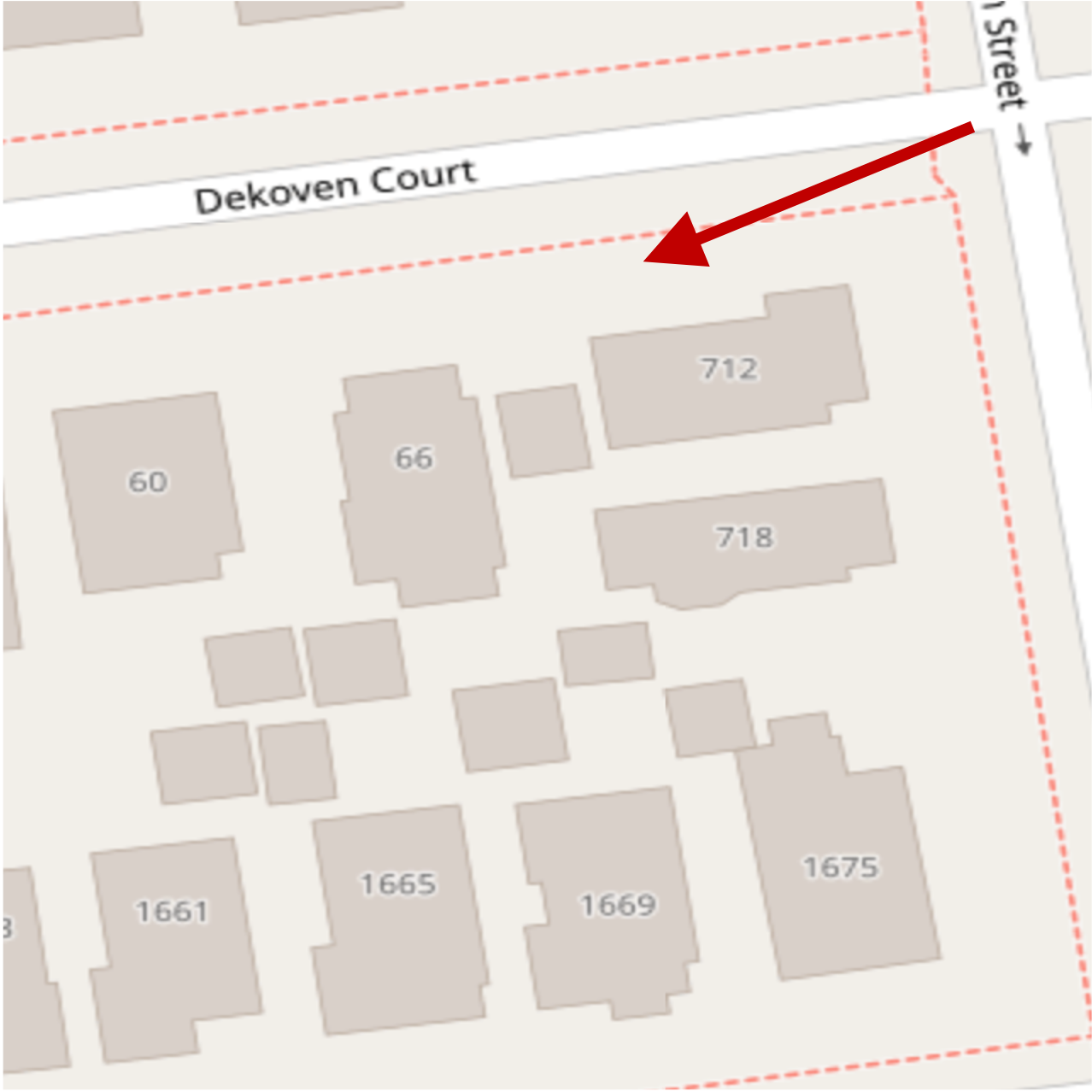} &
         \includegraphics[height=2.7cm]{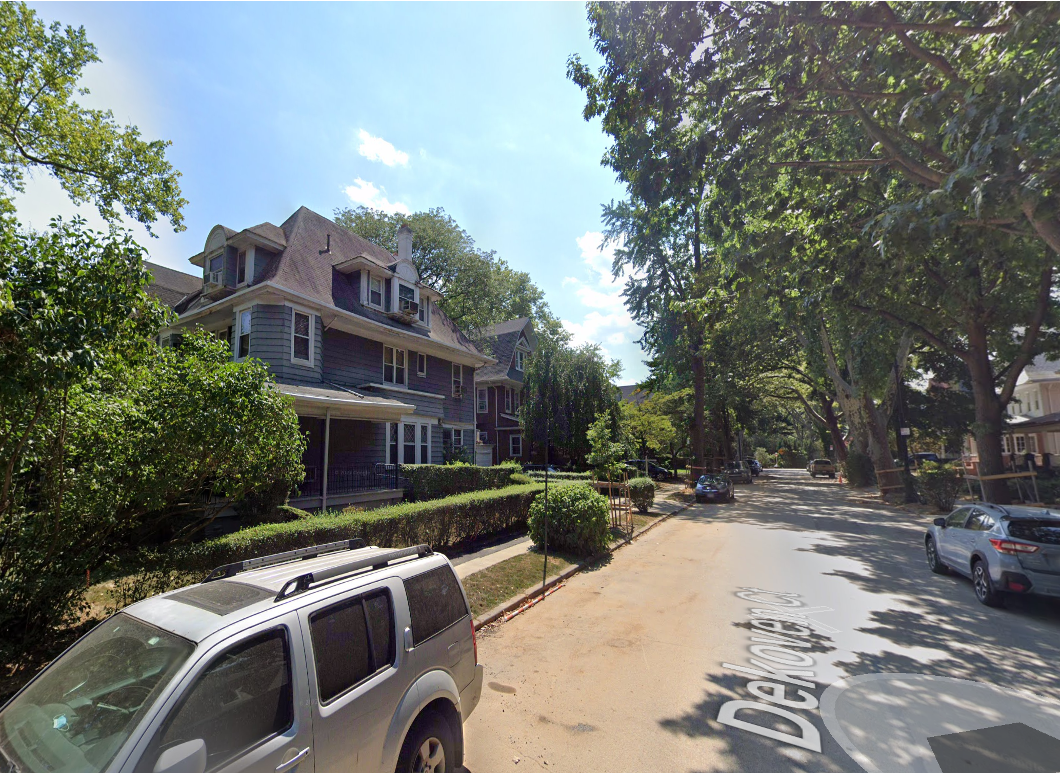} \\
         (a) optical image & (b) elevation (m) & (c) land cover & (d) OSM & (e) street view
    \end{tabular}
    \caption{An example of a potentially incorrectly annotated LCZ 4 patch with corresponding (d) OpenStreetMap data and (e) Google street view photo. The red arrow in (d) points towards to the direction from which the street view photo was taken.}
    \label{fig:failexample}
\end{figure*}

\subsection{Rule-Based LCZ Parameters} \label{sec:exp:acc}

For LCZ mapping $z=z(\mathbf x)$ with parameter vector $\mathbf x(\mathbf y,\mathbf h)=[\text{BSF}(\mathbf y),\text{ISF}(\mathbf y),\text{PSF}(\mathbf y),\text{HRE}(\mathbf y,\mathbf h)]^T$ such that $z=z(\mathbf y, \mathbf h)$, we apply the two sets of thresholds from \cref{tab:thr} to the SFs and HRE: one with ground truth land cover masks $\mathbf y$, and one with their noisy counterparts $\mathbf y=\mathbf y(\mathbf h)$ from \cite{albrecht_autogeolabel_2021}\footnote{In case of noisy labels: BSF, ISF, and PSF utilize the \textit{Buildings}, \textit{Roads}, and \textit{others}\slash\textit{background} classes for $\mathbb L_\text{SF}$, respectively.}.
Since we restrict our study to four parameters characterizing eight LCZs, we recast the LCZ--classification as a multi-label task: $\hat Z(\mathbf y,\mathbf h)=\left\{\hat z:~x_j(\mathbf y,\mathbf h)\in\Delta^{(\hat z)}_j,~j=1\dots4\right\}$. Our performance evaluation is based on whether or not $\hat Z(\mathbf y,\mathbf h)$ covers the ground truth, $z\in\hat Z$ or $z\notin\hat Z$. The class-wise results ($z$ fixed) are summarized in \cref{tab:acc}. Two insights we extract:
\begin{itemize}[leftmargin=3ex,topsep=0pt,itemsep=0ex,partopsep=0ex,parsep=0ex]
    \item Data-adjusted intervals $\Delta$ for the SFs and HRE significantly improve the classification performance, which suggests that the projection between parameters and LCZs requires further inspection and modification---in particular for LCZs 4, 8, and 10.
    \item Despite the application of noisy land cover masks leading to degraded classification results, the low-cost label generation combined with the improvement gained by estimating $\Delta$ from a small subset of accurate labels $z$ indicates the potential of our approach for efficient LCZ mapping.
\end{itemize}

\begin{table}[t]
\scriptsize
\caption{Accuracy assessment (in percent) of LCZ mapping results obtained by \textit{AutoLCZ} with two sets of thresholds from \cref{tab:thr} employing ground truth land cover masks and noisy land cover masks, respectively.}
\label{tab:acc}
\begin{tabular}{m{1.cm}<{\centering}|m{1.2cm}<{\centering}m{1.5cm}<{\centering}|m{1.2cm}<{\centering}m{1.5cm}<{\centering}}
\hline\hline
& \multicolumn{2}{c}{\textbf{Using GT labels}}     & \multicolumn{2}{c}{\textbf{Using noisy labels}}  \\
\cline{2-5}
$z=\text{LCZ}$       & \textbf{Given thresholds} & \textbf{Estimated thresholds} & \textbf{Given thresholds} & \textbf{Estimated thresholds} \\
\hline\hline
1                      & 43.02            & 81.12 {\color[HTML]{3531FF}(+38.10)}                & 56.73            & 69.13 {\color[HTML]{3531FF}(+12.40)}              \\
2                      & 48.58            & 54.78 {\color[HTML]{3531FF}(+\;\;6.20)}             & 69.11            & 61.30 {\color[HTML]{3531FF}(-\;\;7.81)}                \\
3                      & 30.56            & 44.22 {\color[HTML]{3531FF}(+13.66)}                & 11.07            & 30.56 {\color[HTML]{3531FF}(+19.49)}               \\
4                      & 8.54             & 56.66 {\color[HTML]{3531FF}(+48.12)}                & 7.35             & 59.36 {\color[HTML]{3531FF}(+52.01)}                \\
5                      & 34.05            & 50.48 {\color[HTML]{3531FF}(+16.43)}                & 39.52            & 40.00 {\color[HTML]{3531FF}(+\;\;0.48)}                \\
6                      & 60.81            & 72.42 {\color[HTML]{3531FF}(+11.61)}                & 21.18            & 35.36 {\color[HTML]{3531FF}(+14.18)}                \\
8                      & 26.84            & 82.89 {\color[HTML]{3531FF}(+56.05)}                & 20.26            & 86.05 {\color[HTML]{3531FF}(+65.79)}                \\
10                     & 1.41             & 82.13 {\color[HTML]{3531FF}(+80.72)}                & 6.11             & 83.23 {\color[HTML]{3531FF}(+77.12)}                \\
\hline
All (OA)               & 44.85            & 59.11 {\color[HTML]{3531FF}(+14.26)}                & 38.91            & 48.56 {\color[HTML]{3531FF}(+\;\;9.65)}      \\
\hline\hline
\end{tabular}
\end{table}

\subsection{LCZ Misclassified vs.\ Mislabeled}

As \cref{tab:acc} demonstrates, the accuracy to predict LCZ classes using pre-defined thresholds may be quite low, cf.\ LCZs 4 and 10. To shed light on potential root causes, we showcase an example of LCZ 4 along with the street view photo retrieved from Google in \cref{fig:failexample}. The street view indicates a three-story building to the left. As a rough proxy we assume 3 meters per story. Hence, the building raises up to about 10m above ground. This height is obviously below the HRE lower bound value of 25m for LCZ 4. The corresponding patch $i$ is likely mislabeled by the (human) annotation process. Such bias in data defining $\Delta$ can lead to low LCZ classification accuracy for our data-driven approach.

\section{Conclusions} \label{sec:conc}

We proposed the \textit{AutoLCZ} framework for boosting the automation of LCZ mapping with the aid of rules that scalably operate on LiDAR data to model LCZ parameters. We applied a proof-of-concept experiment for New York City on four LCZ parameters, namely three surface fraction measures and the Roughness Element (of height). The results indicate that \textit{AutoLCZ} is promising in not only large-scale LCZ mapping, but also it proofs useful in the correction and quality control of manually annotated LCZ labels. Our initial findings motivate to further improve the accuracy and efficiency of \textit{AutoLCZ} by refining the set of rules for LCZ parameter estimation. Exploring more efficient classification strategies with the aid of machine learning techniques is a natural follow-up to our experiments. Integration of thermal modalities (e.g., band 10 of the Landsat 8 satellite) may provide an additional boost in performance to the \textit{AutoLCZ} methodology. 

\bibliographystyle{IEEEbib}

\end{document}